\documentclass{article}

\usepackage{microtype}
\usepackage{graphicx}
\usepackage{subfigure}
\usepackage{booktabs} 

\usepackage{hyperref}

\usepackage[accepted]{icml2023}

\usepackage{amsmath}
\usepackage{amssymb}
\usepackage{mathtools}
\usepackage{amsthm}

\usepackage[capitalize,noabbrev]{cleveref}

\theoremstyle{plain}

\theoremstyle{definition}

\theoremstyle{remark}

\usepackage[textsize=tiny]{todonotes}

\icmltitlerunning{Invalid Logic, Equivalent Gains: The Bizarreness of Reasoning in Language Model Prompting}

\begin{document}

\twocolumn[
\icmltitle{Invalid Logic, Equivalent Gains:\\The Bizarreness of Reasoning in Language Model Prompting}



\icmlsetsymbol{equal}{*}

\begin{icmlauthorlist}
\icmlauthor{Rylan Schaeffer}{equal,stanfordcs}
\icmlauthor{Kateryna Pistunova}{equal,stanfordphysics}
\icmlauthor{Samar Khanna}{equal,stanfordcs}
\icmlauthor{Sarthak Consul}{equal,stanfordcs}
\icmlauthor{Sanmi Koyejo}{stanfordcs}
\end{icmlauthorlist}

\icmlaffiliation{stanfordcs}{Department of Computer Science, Stanford University}
\icmlaffiliation{stanfordphysics}{Department of Physics, Stanford University}

\icmlcorrespondingauthor{Rylan Schaeffer}{rschaef@cs.stanford.edu}
\icmlcorrespondingauthor{Sanmi Koyejo}{sanmi@cs.stanford.edu}

\icmlkeywords{Reasoning, Large Language Models, Language Models, ICML, Prompting, Chain-of-Thought}

\vskip 0.3in
]



\printAffiliationsAndNotice{\icmlEqualContribution} 

\begin{abstract}
Language models can be prompted to reason through problems in a manner that significantly improves performance.
However, \textit{why} such prompting improves performance is unclear. Recent work showed that using logically \textit{invalid} Chain-of-Thought (CoT) prompting improves performance almost as much as logically \textit{valid} CoT prompting, and that editing CoT prompts to replace problem-specific information with abstract information or out-of-distribution information typically doesn't harm performance.
Critics have responded that these findings are based on too few and too easily solved tasks to draw meaningful conclusions.
To resolve this dispute, we test whether logically invalid CoT prompts offer the same level of performance gains as logically valid prompts on the hardest tasks in the BIG-Bench benchmark, termed BIG-Bench Hard (BBH).
We find that the logically \textit{invalid} reasoning prompts do indeed achieve similar performance gains on BBH tasks as logically valid reasoning prompts.
We also discover that some CoT prompts used by previous works contain logical errors. 
This suggests that covariates beyond logically valid reasoning are responsible for performance improvements.
\end{abstract}

\section{Introduction}
\label{sec:introduction}

Language models can perform significantly better when prompted in particular ways.
For example, prompts that recommend or guide language models through step-by-step processing have been shown to significantly improve performance on question answering, conversational response generation and other tasks \cite{nye2021show, wei2022chain,jung2022maieutic, kojima2022large,yao2023tree}.
These prompting techniques are especially effective on the hardest tasks \cite{suzgun2022challenging} in the BIG-Bench benchmark \cite{srivastava2022beyond}, leading many to conclude that such techniques unlock emergent\footnote{Although see \citet{schaeffer2023emergent}.} human-like reasoning abilities in large language models \cite{wei2022emergent}. However, \textit{why} such prompting strategies improve performance is unclear. 
\citet{madaan2022text} showed replacing problem-specific information in Chain-of-Thought (CoT) prompts with either abstract information or out-of-distribution information typically doesn't harm CoT's performance gains, and \citet{wang2022towards} showed that logically \textit{invalid} CoT prompts (i.e., prompts which contain logically invalid chains of reasoning) achieve approximately the same gains as logically valid CoT prompts.
These discoveries are puzzling, but critics warned against drawing confident or far-ranging conclusions results because (1) only a small number of tasks are considered (specifically GSM8K \cite{cobbe2021training}, Bamboogle \cite{press2022measuring}, and two commonsense tasks \cite{srivastava2022beyond}) and (2) the tasks are relatively easy to solve.

In this work, we aim to clarify this dispute by asking whether logically invalid CoT prompts indeed offer comparable performance gains as logically valid CoT prompts in \textit{more} and \textit{harder} tasks. To answer this question, we evaluate the performance of logically invalid CoT prompts on the hardest tasks in BIG-Bench \cite{srivastava2022beyond}, the so-called BIG-Bench Hard (BBH) tasks \cite{suzgun2022challenging}.
We find that logically invalid CoT prompts induce performance gains approaching logically valid CoT prompts on BBH tasks.
We also discover that CoT prompts from previous work contain logical errors, a discovery that we confirmed with the original authors.
Our findings support growing evidence that while prompting techniques can, without doubt, significantly improve language model performance on complex tasks, there is questionable evidence that these performance gains are attributable to logical reasoning skills.
Covariates in the prompts other than logical reasoning may be responsible for the performance improvements and point to the need for continued investigation. 

\section{Methods}
\label{sec:method}

\subsection{Tasks: BIG-Bench Hard $\subset$ BIG-Bench}
\label{sec:method:dataset}

Beyond-the-Imitation Game Benchmark (BIG-Bench) is a benchmark of over 200 natural language tasks created to evaluate the capabilities and limitations of language models \cite{srivastava2022beyond}.
BIG-Bench Hard (BBH) is a subset of BIG-Bench consisting of 23 tasks specifically identified as extremely difficult for language models \cite{suzgun2022challenging}.
BBH was constructed by excluding BIG-Bench tasks using the following criteria: tasks with more than three subtasks, tasks with fewer than 103 examples, tasks without human-rater baselines, tasks that do not use multiple-choice or exact match as the evaluation metric, and tasks on which the best reported model beats average reported human-rater score; the remaining 23 BIG-Bench tasks comprise the BBH tasks.
The BBH tasks cover a variety of categories, including traditional NLP, mathematics, commonsense reasoning, and question-answering.
In general, BBH tasks typically fall into one of two categories: more traditional NLP tasks (e.g., Date Understanding) or more algorithmic tasks (e.g., Boolean Expressions).
Many state of the art language models, including GPT-3 \cite{brown2020language} and PaLM \cite{chowdhery2022palm}, as well as internal dense and sparse Google models, were shown to be incapable of solving BBH tasks above average human rater performance when asked directly.

\subsection{Prompt Types}
\label{sec:method:prompting}

In this work, we compare three prompt types applied to the BBH tasks. Examples of task problems and prompts are shown in Table \ref{tab:prompt_examples}.

\paragraph{Chain-of-Thought (CoT) Prompting}

In (logically valid) Chain-of-Thought (CoT) prompting, each question in each task is prepended with three high-quality, human hand-written, topical examples of Question$\rightarrow$Reasoning$\rightarrow$Answer.
The reasoning is a sequence of natural language steps that derive the answer from information in the question \cite{nye2021show, wei2022chain}.
For each BBH task, \citet{suzgun2022challenging} released three examples of Question$\rightarrow$Reasoning$\rightarrow$Answer that we adapt. We do not use their prompts directly because we want CoT prompts to contain logically valid chains of reasoning, and as we discovered (Sec. \ref{sec:results:errors}), some of Suzgun et al.'s CoT prompts contain logical errors.

\paragraph{Logically Invalid Chain-of-Thought (Invalid CoT) Prompting}

In logically invalid Chain-of-Thought (Invalid CoT) prompting, we edit the reasoning in the three examples for all three tasks to become logically invalid. To do this, we modify the reasoning to contain nonsensical or invalid steps. For instance, on the Boolean Expressions task, we provided the reasoning, ``Because English does not permit multiple negatives, the expression "(not not True)" evaluates to "( not True )", and on the Date Understanding task, we provided the reasoning, ``If tomorrow is 11/12/2019, then today is 11/12/2018. The date one year ago from today is 11/11/2018." In all cases, the modified BBH exemplars were logically invalid, but reached the correct answers. We wrote each example once and did not modify the examples whatsoever to improve task performance.

\paragraph{Answer-Only (AO) Prompting}

To establish baseline performance, we follow previous work and prompt language models to answer each question in each task directly. We call this ``Answer Only" (AO) prompting.

\begin{table}[htbp]
    \centering
    \begin{tabular}{p{0.25\linewidth}p{0.65\linewidth}}
        \toprule
        \textbf{Prompt Type} & \textbf{Example Query} \newline\newline Evaluate the result of a random Boolean expression. \newline\newline
        Q: not ( ( not not True ) ) is \\
        \midrule
        Answer Only & False\\
        \midrule
        Scratchpad/Chain-of-Thought (CoT) & A: Let's think step by step.
        Remember that (i) expressions inside brackets are always evaluated first and that (ii) the order of operations from highest priority to lowest priority is "not", "and", "or", respectively.
        We first simplify this expression "Z" as follows: "Z = not ( ( not not True ) ) = not ( ( A ) )" where "A = not not True".
        Let's evaluate A: A = not not True = not (not True) = not False = True.
        Plugging in A, we get: Z = not ( ( A ) ) = not ( ( True ) ) = not True = False. So the answer is False.\\
        \midrule
        Logically-Invalid Chain-of-Thought (Invalid CoT) & A: Let's think step by step.
        Remember that (i) expressions inside brackets are always evaluated first and that (ii) the order of operations from highest priority to lowest priority is "not", "and", "or", respectively.
        Because English does not permit multiple negatives, the expression "(not not True)" evaluates to "( not True )".
        The expression "not ( ( not not True ) )" therefore evaluates to "not ( ( not True ) )".
        By the same logic, the expression "not ( ( not True ) )" simplifies to "not True". In Boolean logic, "not True" is False. So the answer is False.\\
        \bottomrule
    \end{tabular}
    \caption{\textbf{Examples of different prompt types.}}
    \label{tab:prompt_examples}
\end{table}





\subsection{Metrics}
\label{sec:method:metrics}

To evaluate the performance of different prompting types, we used Accuracy (i.e. Exact String Match) to evaluate model outputs. We acknowledge that recent work showed such a choice of metric may reach misleading scientific conclusions \cite{schaeffer2023emergent}, but made the choice to ensure fair comparison with the relevant preceding work.

\subsection{Choice of Language Model}
\label{sec:method:model}

\begin{figure}
    \centering
    \includegraphics[width=0.9\linewidth]{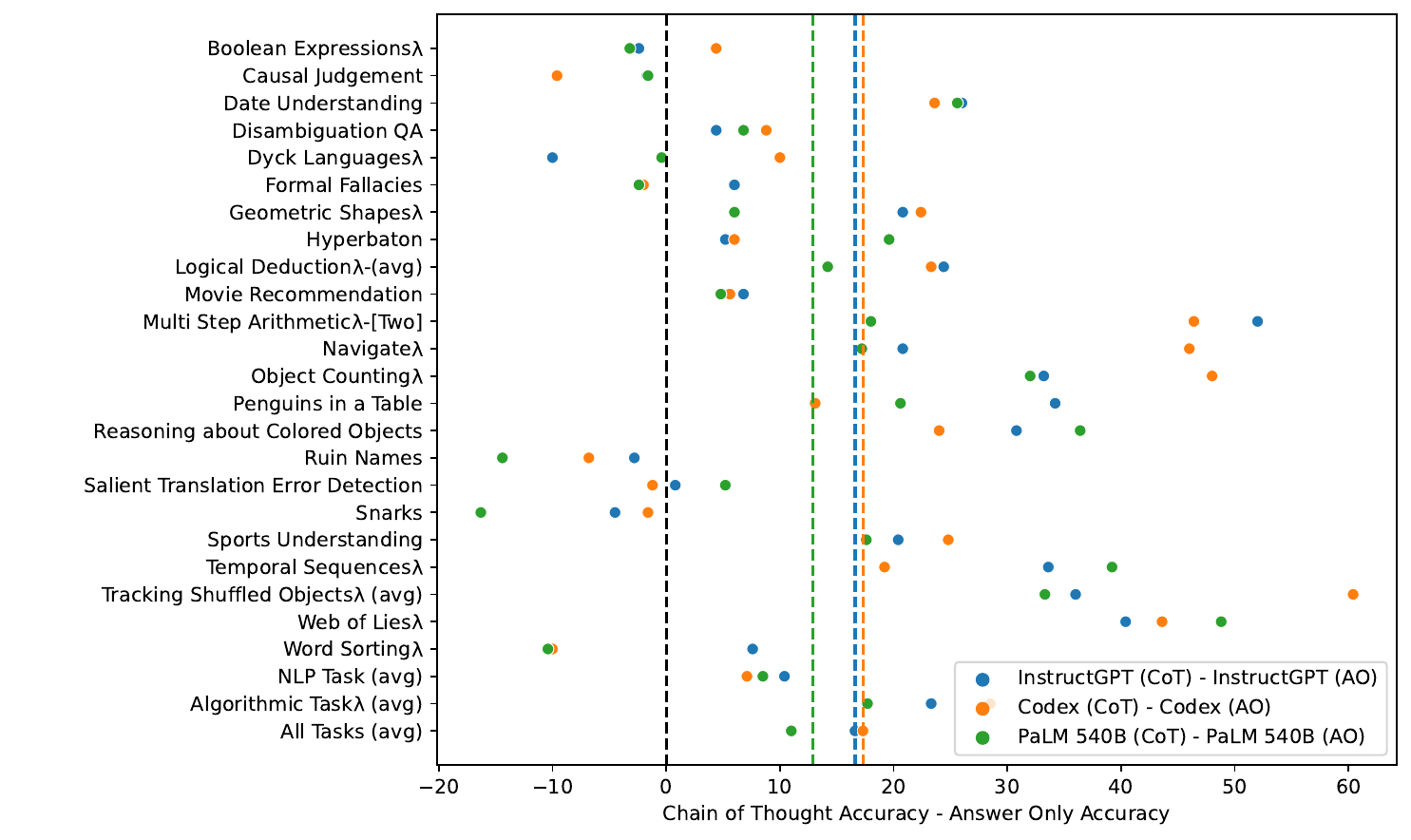}
    \caption{\textbf{Codex benefits more from Chain-of-Thought Prompting than InstructGPT or PaLM 540B on BIG-Bench Hard Tasks.} Dashed vertical lines indicate average model performance across all BIG-Bench Hard Tasks. Data from \citet{suzgun2022challenging}.}
    \label{fig:suzgun_cot_minus_human_avg}
    \vspace{-6pt}
\end{figure}

Due to limited resources, we could only evaluate one model.
Only Codex \cite{chen2021evaluating} and InstructGPT \cite{ouyang2022training} were publicly queryable, thus ruling out PaLM 540B \cite{chowdhery2022palm}.
We chose to evaluate Codex because our visualizing Suzgun et al's data revealed Codex outperformed both InstructGPT and PaLM 540B; this point was not made by the authors, but visualizing their data anew reveals this conclusion (Fig. \ref{fig:suzgun_cot_minus_human_avg}).
Although this finding may seem surprising, independent work has found that language models reason better if pretrained heavily on code \cite{madaan2022language}, as Codex was.
Perhaps unsurprisingly, we found that Codex's high average performance with CoT prompting is driven by large improvements on BBH algorithmic tasks, overriding mediocre improvements (or regressions) on natural language tasks.

\subsection{Choice of BBH Tasks}
We evaluate AO, CoT, and Invalid CoT prompting strategies on all BBH tasks.

\section{Experimental Results}
\label{sec:results}

\begin{figure}
    \centering
    \includegraphics[width=0.9\linewidth]{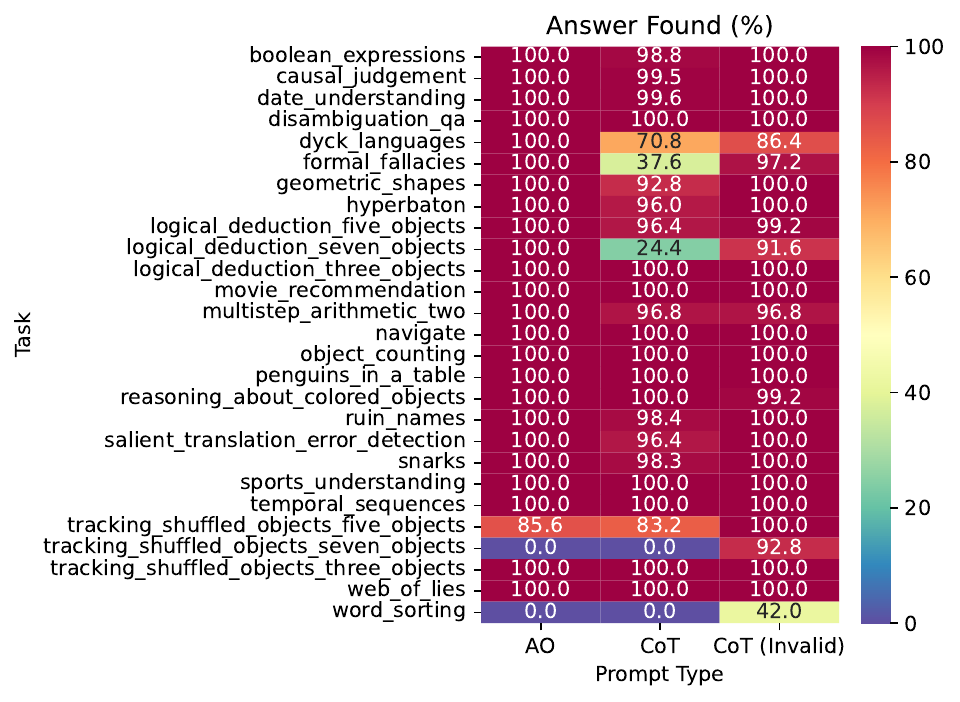}
    \includegraphics[width=0.45\linewidth]{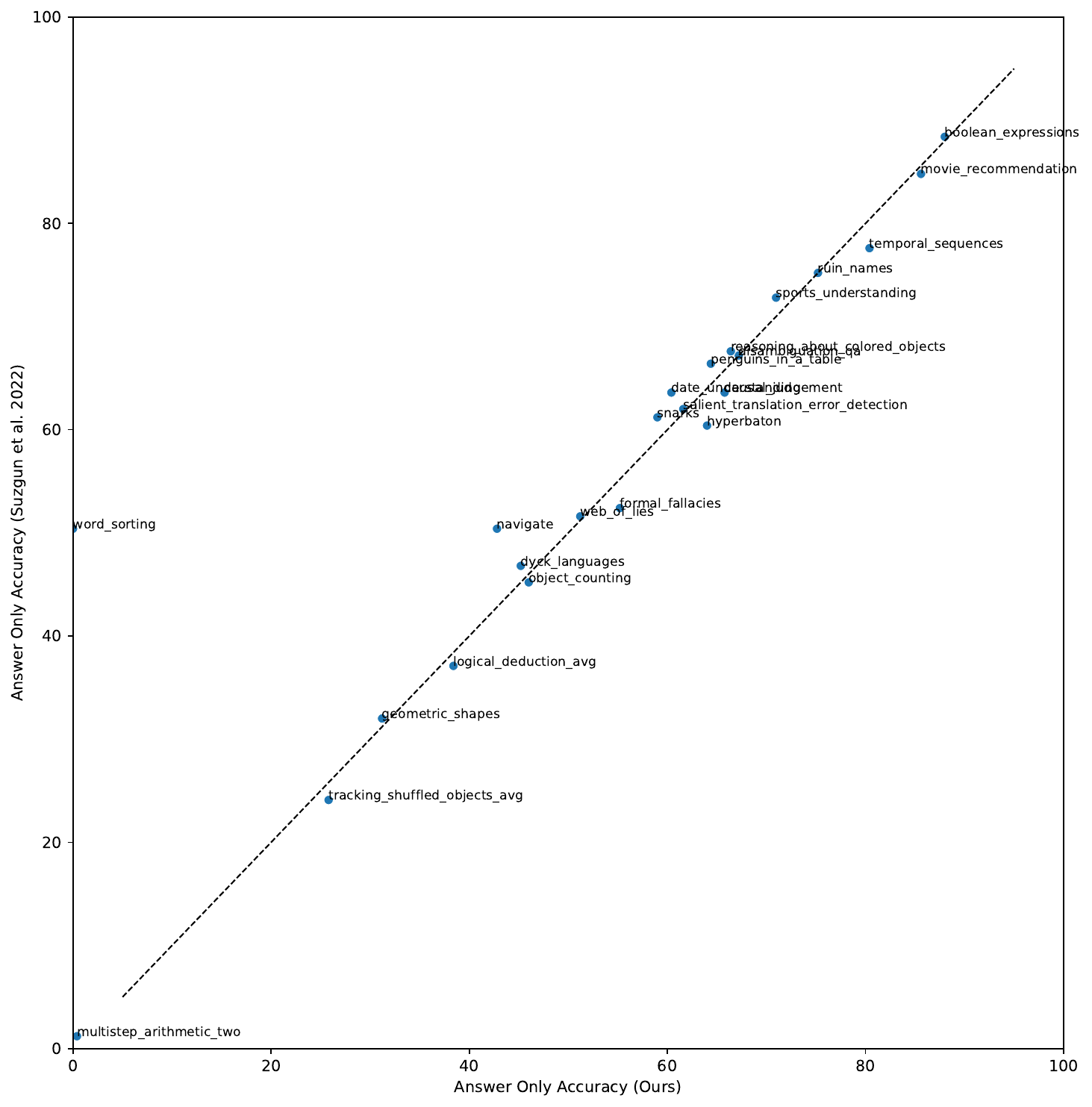}%
    \includegraphics[width=0.45\linewidth]{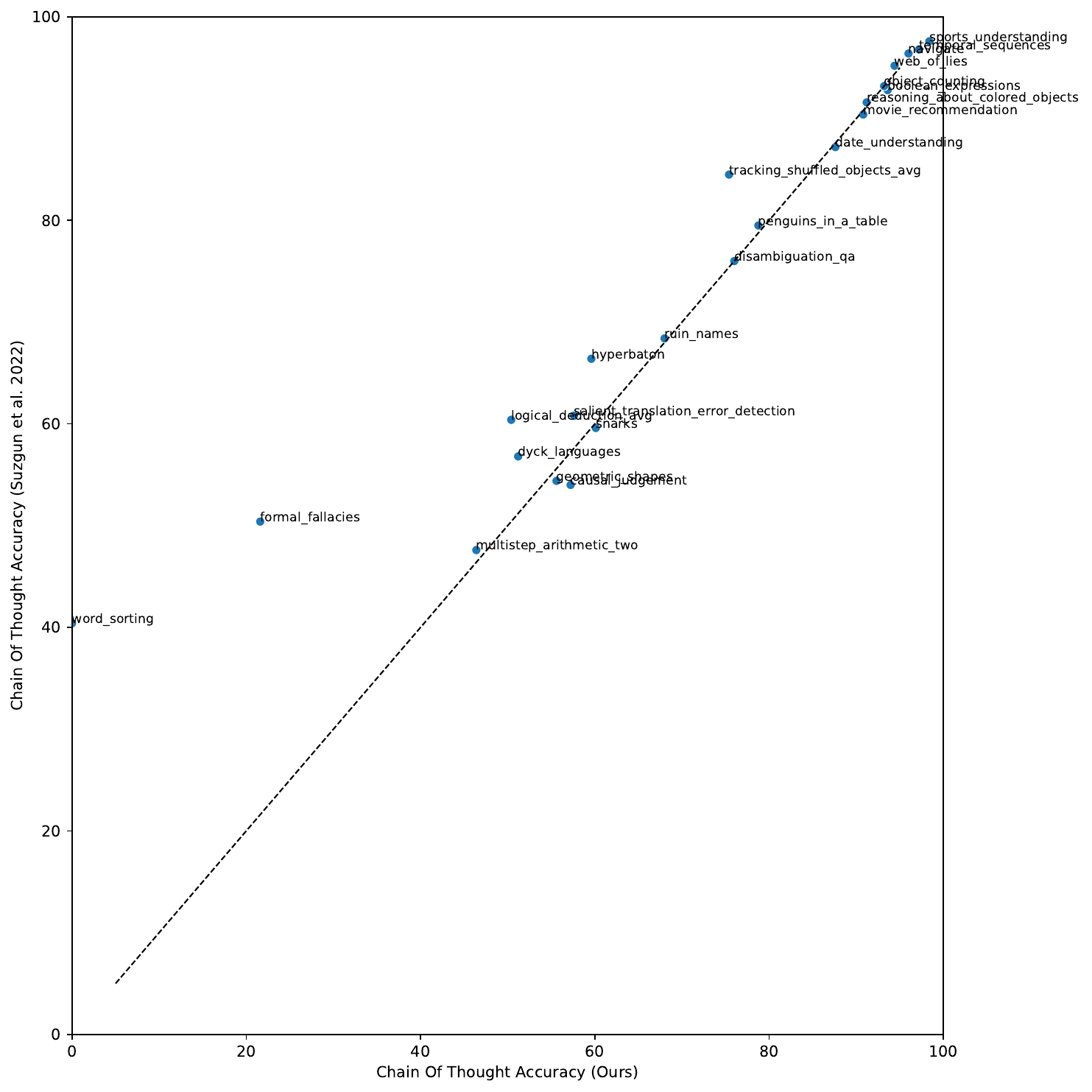}
    \caption{\textbf{Sanity checks.} Top: All prompting strategies almost always produced an answer across tasks. Bottom: Our accuracies for Answer Only (AO) and Chain-of-Thought (CoT) prompting closely matched published values from \citet{suzgun2022challenging} except on one task: word\_sorting.}
    \label{fig:sanity_check}
    \vspace{-6pt}
\end{figure}

\subsection{Previous CoT Prompts Contain Logical Errors}
\label{sec:results:errors}

While converting the logically valid CoT prompts written for BBH tasks by \citet{suzgun2022challenging} to logically invalid CoT prompts, we discovered that at least three of the BBH tasks' CoT prompts were \textit{already} logically invalid.
On the Multistep Arithmetic Task, a human handwritten CoT prompt reads: ``Then, the final equation is A * B = -41 * -3 = (-61) * (-3) = 123. So the answer is 123."
To clarify, -41 was substituted with -61 without justification, and -61 * -3 is not 123, even though 123 is the correct answer.
On the Navigate task, a human handwritten CoT prompt reads: ``(4) Take 7 steps right: (0, 7), facing the positive y-axis." should be ``(4) Take 7 steps right: (0, 0), facing the positive y-axis."
On the Web of Lies task, ``(3) Vina says Jerry tells the truth. Since we know from (2) that Jerry tells the truth, if Vine says Jerry tells the truth, then Vina tells the truth." should be ``(3) Vina says Jerry tells the truth. Since we know from (2) that Jerry tells the truth, if Vina says Jerry tells the truth, then Vina tells the truth."
All three errors were raised to the authors, who confirmed our discoveries and accepted our suggested corrections as well.
This discovery already hints that CoT prompts need not be logically correct for the model to output the correct answer.
It also hints that previous CoT prompts from earlier papers, e.g., \cite{wei2022chain}, may need to be investigated further.

\subsection{Sanity Checks}

We found that Codex produced an answer under all prompting strategies on almost all tasks (Fig. \ref{fig:sanity_check} top), and the accuracies under Answer Only (AO) and Chain-of-Thought (CoT) prompting closely matched published values \cite{suzgun2022challenging}. The one exception was a single task: word\_sorting.

\begin{figure}
    \centering
    \includegraphics[width=0.9\linewidth]{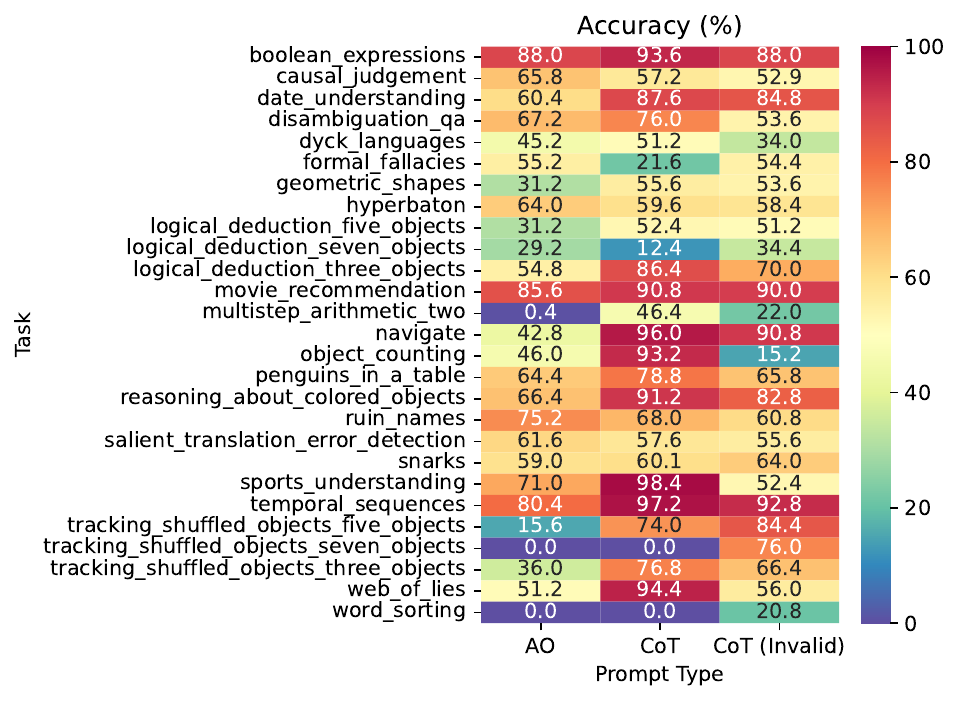}
    \caption{\textbf{Accuracy per task per prompt type on BIG-Bench Hard.} Prompt Types: AO = Answer Only, CoT = Chain-of-Thought, CoT (Invalid) = Logically Invalid Chain-of-Thought. Approximately 200-250 questions per BIG-Bench Hard task.}
    \label{fig:accuracies_per_task}
    \vspace{-6pt}
\end{figure}

\subsection{Accuracy by Prompt Type by Task}

\begin{figure}[t]
    \centering
    \includegraphics[width=0.9\linewidth]{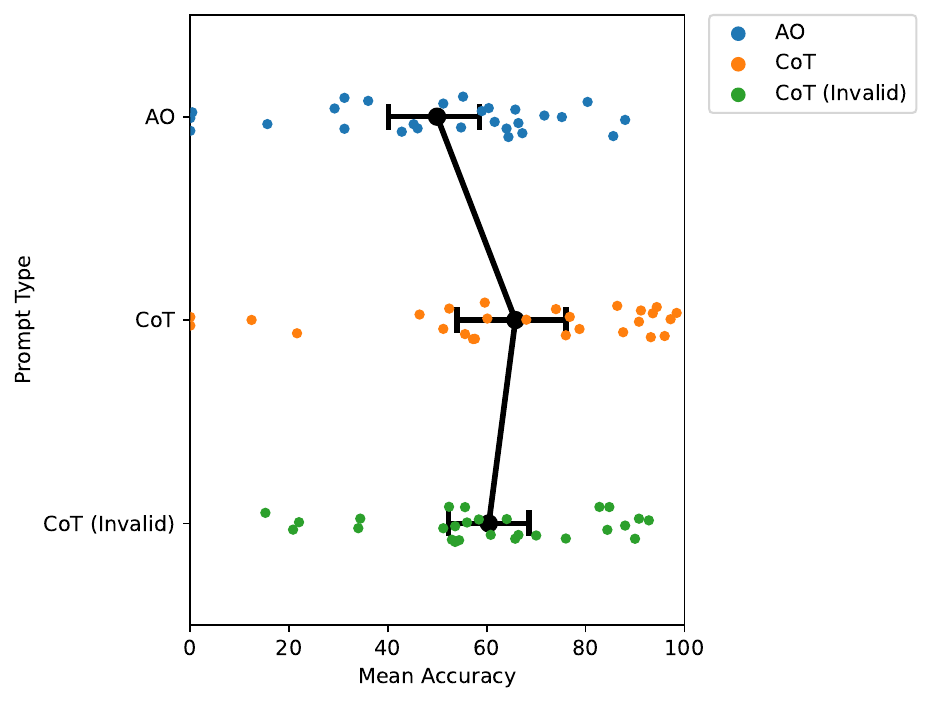}
    \includegraphics[width=0.9\linewidth]{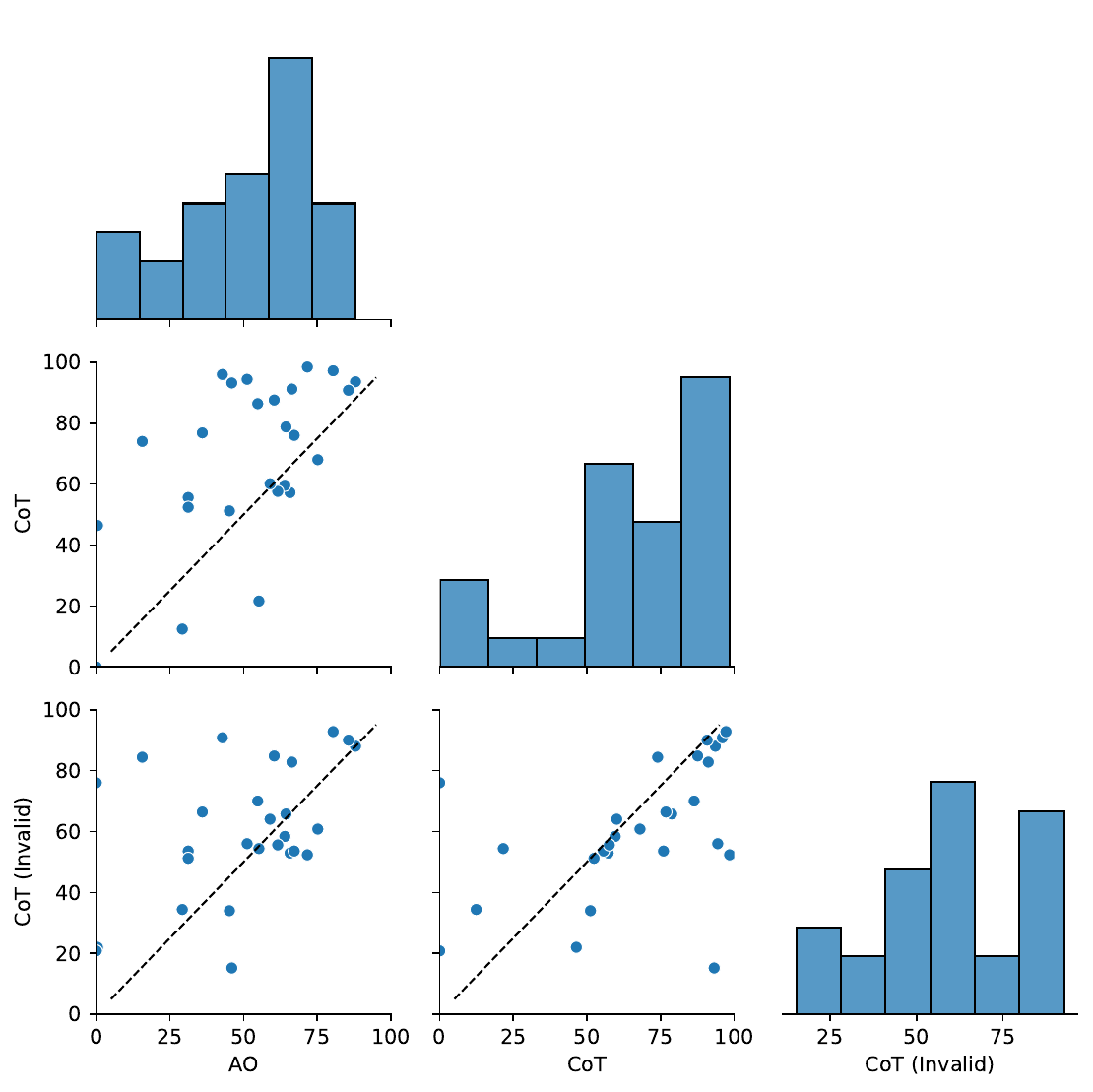}
    \caption{\textbf{Logically invalid Chain-of-Thought outperforms Answer Only and nearly matches Chain-of-Thought.} Top: Accuracy per BBH task, averaged over all BBH tasks. Bottom: Pairwise accuracy plots comparing the three different prompt types.}
    \label{fig:accuracies_analysis}
    \vspace{-6pt}
\end{figure}

The Accuracy of Codex on each of the BIG-Bench Hard (BBH) tasks under each of the three prompt types (Answer Only, Chain-of-Thought, Logically Invalid Chain-of-Thought) is displayed in Fig. \ref{fig:accuracies_per_task}. Despite using the same model and the same decoding hyperparameters (e.g., temperature), we notice that there is tremendous variation in the accuracies. To better compare the performance across the three prompt types, we visualized each prompt type's average accuracy over all BBH tasks. We found that Chain-of-Thought prompting beats Answer Only prompting, but Logically Invalid Chain-of-Thought is close behind Chain-of-Thought and better than Answer Only (Fig. \ref{fig:accuracies_analysis} top). Pairwise comparisons between prompt types across different BBH tasks reveal that different prompt types do not dominate one another (Fig. \ref{fig:accuracies_analysis} bottom), but rather display rich structure. We note that we do not optimize logically invalid CoT prompts whatsoever, meaning the Invalid CoT performance here is likely below what tuning could achieve.

\section{Conclusion}

On the diverse and challenging BIG-Bench Hard tasks, we find that Chain-of-Thought prompting performs best on average, but logically invalid Chain-of-Thought prompting is close behind and outperforms Answer Only prompting. This demonstrates that completely illogical reasoning in the CoT prompts do not significantly harm the performance of the language model. Our findings suggest that valid reasoning in prompting is not the chief driver of performance gains, raising the question of what is.
We note that there are complementary approaches towards achieving reasoning in language models such as enforcing valid reasoning in the model architecture \cite{creswell2022selection, creswell2022faithful} or via autoformalization \cite{wu2022autoformalization, azerbayev2023proofnet}.

Our work raises important questions for future work. Why are models robust to invalid CoT prompts? What features of the data or prompts result in the model outputting inconsistent or invalid outputs?
Does increasing the degree of "incorrectness" or the number of incorrect prompts affect the model's sensitivity to invalid CoT?
What other properties of the valid prompts is the model sensitive to? Answering these questions can yield useful insights into prompt engineering for LLMs, as well as a deeper understanding of when models output inconsistent or ``hallucinated" answers.



\clearpage

\bibliography{references.bib}
\bibliographystyle{icml2023}

\clearpage

\end{document}